# A Semantic Approach for Automatic Structuring and Analysis of Software Process Patterns

Nahla JLAIEL
RIADI Research Laboratory
National School of Computer Science
La Manouba 2010, Tunisia

Khouloud Madhbouh
Higher Institute of Multimedia and Computer Science
University campus Erriadh city
6072 Zirig Gabès, Tunisia

Mohamed BEN AHMED
RIADI Research Laboratory
National School of Computer Science
La Manouba 2010, Tunisia

## ABSTRACT
The main contribution of this paper, is to propose a novel semantic approach based on a Natural Language Processing technique in order to ensure a semantic unification of unstructured process patterns which are expressed not only in different formats but also, in different forms. This approach is implemented using the GATE text engineering framework and then evaluated leading up to high-quality results motivating us to continue in this direction.

## 1. INTRODUCTION
Software process patterns are being considered as a valuable mechanism to capture and disseminate best practices during software development processes. Consequently, they have been successfully and increasingly used within software development communities to reuse proven solutions.

In this context, many formalisms and languages have been proposed to describe software process patterns. This multiplicity makes capitalization and/or reuse of process patterns, difficult to be achieved [1].

In this paper, we propose a semantic approach named ASAP acronym for ''Automatic Structuring and Analysis of process Patterns'' based on a linguistic method of Natural Language Processing (NLP) in order to provide architectural and semantic unification of unstructured patterns which are described in different formats (e.g. PDF, WORD, HTML, etc.) and different forms (e.g. Ambler, Störrle, PPDL, UML-PP, PPL, etc.) using the GATE API. The remainder of this paper is organized into six sections. Section 2 introduces the context of our research work. Section 3 provides background information on process patterns unification and the NLP methodology. Section 4 describes the proposed approach named ASAP. Section 5 details the experimentations results of the proposed approach. Section 6 concludes the paper by giving a discussion of our contibiution as well as an overview of our future work.

## 2. PROCESS PATTERNS REUSE
Patterns are increasingly being recognized by software development communities, as an effective method to reuse knowledge and best practices gained during software development processes [2] [3]. Indeed, they are growing to be widely used as proven solutions to recurring problems consisting essentially of a triplet of problem, context and solution. In addition, patterns are not restricted to a particular domain to be applied in or to emerge of. Instead, they have been developed for several domains e.g. Architecture, Software Engineering, Organization, Pedagogy as well as Human Computer Interaction.

Consequently, software patterns nowadays exist for a wide range of topics including requirement patterns, analysis, design, implementation or code patterns, test patterns and even maintenance patterns.

Most of these latter consist of product or result patterns whose role is to capitalize specifications or implementations of a goal. Concerning process patterns, whose main role is to capitalize good specifications or implementations of a method to be followed to achieve a goal [4], they become commonly used by software development communities as an excellent medium to share software development knowledge that is often encapsulated in experiences and best practices [1] [5] [6] [7].

Indeed, process patterns are growingly being adopted by different development processes such as Agile processes [8], Object-oriented Software Development processes [9], Component Based Software Development processes [10], Service-Oriented Development processes [11] as well as Aspect-oriented Development processes [12].

As consequence to the huge proliferation of the process patterns practice, these latter are being used in an informal manner, through traditional textbooks or better with modest hypertext systems providing weak semantic relationships. In addition to the huge number of process patterns that are available in books or Web-based resources [3], they significantly differ in format, coverage, scope, architecture and terminology used [1].

All of these observations conspire to create barriers to the efficient use and reuse of process patterns.

In fact, patterns users are expected to investigate different patterns resources such as books, magazines, papers and Web collections to find the most appropriate patterns. This investigation really needs cognitive efforts, abilities and time to identify, understand, select, adapt and apply relevant ones.

For these reasons, we have argued in a previous work [13], that efforts are needed to more formally capitalize patterns knowledge in order to help software development communities use, reuse and create process patterns during any given software development process.

Consequently, we have set ourselves as an overall goal of research to build up an intelligent framework supporting process patterns capitalization and reuse. For this purpose, we proposed a holistic approach named SCATTER [13], acronym for "SemantiC Approach for sofTware process paTErns capitalization and Reuse", which aims to disseminate software process best practices by making process patterns described in a unified and formal form. The proposed approach is based on two main processes namely, process patterns warehousing and process patterns mining.

The present work takes place into the targeted framework and forms a first part of the overall proposed approach. In this context, we implement a semantic approach for process patterns unification by analyzing and structuring their description in order to facilitate and enhance patterns reuse within software development communities.

## 3. BACKGROUND

This section is intended to provide background information for the proposed approach ASAP. The first subsection is devoted to the description of the process pattern unification model as the building block of the proposed approach. The next subsection deals with the natural language processing method and tools as the adopted methodology in this work.

### 3.1 Process Patterns' Unification

#### 3.1.1 Process Patterns' Reality

Different initiatives have been carried out in the literature of patterns dealing with process patterns' description and formalization. These are classified into description models such as Ambler [9], RHODES[14], Gnatz [15], P-Sigma [4], Störrle [16] and other as languages, such as PROMENADE [17], PPDL [6], PROPEL [18], PLMLx [19], UML-PP [7] and PPL [20].

Several lacks have been revealed from the survey that we carried out in a previous work [21] concerning the aforementioned works, based on eleven evaluation criteria. Detailed in [1] and [21], these latters conspire to create barriers to patterns' knowledge capitalization and reuse. Among these, we notice the lacks of architectural as well as terminological consent in patterns descriptions.

The lack of architectural consent means that different process pattern descriptions have been proposed using disparate architectures. In fact, when comparing the eleven selected works from the literature, we identified eleven different pattern description facets, namely: identification, classification, problem, context, solution, role, artifact, relationship, guidance, management and evaluation [1]. In addition, these are differently covered by process patterns descriptions and most of them pay more attention to the four main facets: context, solution, problem and relationships of a pattern.

The lack of terminological consent refers to the problems of polysemy and synonymy addressed in labels used to describe patterns. Indeed, we find terms such as Consequences used to express a Resulting Context in PPL as well as a Guideline in Gnatz. Moreover, others different terms are being used to address the same concept such as Intention in RHODES to describe a pattern Problem, instead, the term Intent is used in Störrle. [13]

#### 3.1.2 A Unified Description of Process Patterns

To overcome the afore mentioned lacks, a first step was to create a unified conceptualization of process patterns. Thus, mappings efforts [1] were necessary to achieve this goal leading to a process patterns' meta-model unifying patterns knowledge representations. In this latter, we consider a process pattern information description from six facets: [13]

The identification facet encapsulates a set of properties identifying a pattern such as pattern name, author(s), keywords, pattern's classification (type, category, abstraction level, and aspect) as well as pattern origin (project and participants) and pattern artifacts (used and/or produced). The core information is the main pattern facet embodying details about the well-known triplet: problem, context and solution. The relationships facet expresses how a pattern could interact with other patterns (e.g. similar patterns, refinement patterns, subsequent patterns, and anti-patterns). The guidance facet refers to the support level provided by a pattern to be comprehended and used (e.g. known uses, example, literature, illustration, etc.). The evaluation facet provides feedbacks on pattern application (e.g. discussion, confidence, maturity, etc.). The management facet provides general information about a given pattern (e.g. version, creation-date).

Figure 1 illustrates the proposed unified description of process patterns according to the above mentioned facets.

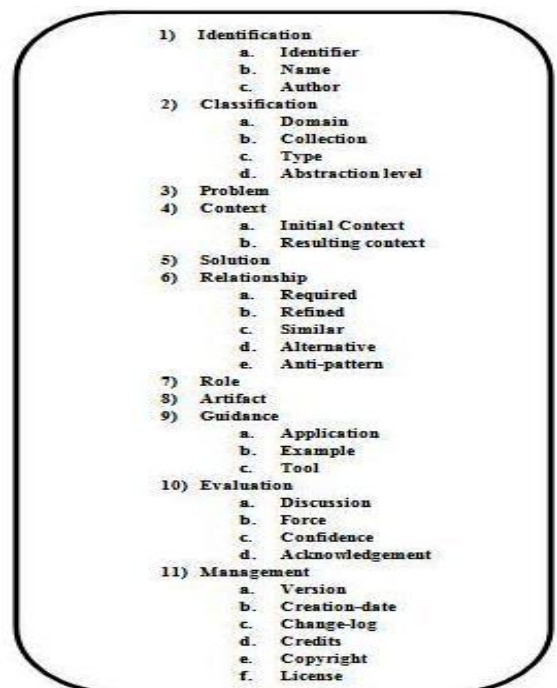

**Figure 1: The adopted unified description of a process pattern**

### 3.2 NLP Methodology

#### 3.2.1 Natural Language Processing

The Natural Language Processing (NLP) is a computerized approach to analyzing language data, expressed in a language called "natural", that is spoken or written.
The NLP is based on both a set of theories and a set of technologies and is being a very active area of research and development. So, there is not a single agreed-upon definition that would satisfy everyone [22].
it aims to model and reproduce, using computers, the human capacity to produce and understand natural languages.

The NLP involves different areas of investigation, namely: Computer Science, Linguistic, Mathematics and Artificial Intelligence.

Figure 3 reveals the main tasks that are commonly included in most of the NLP applications:

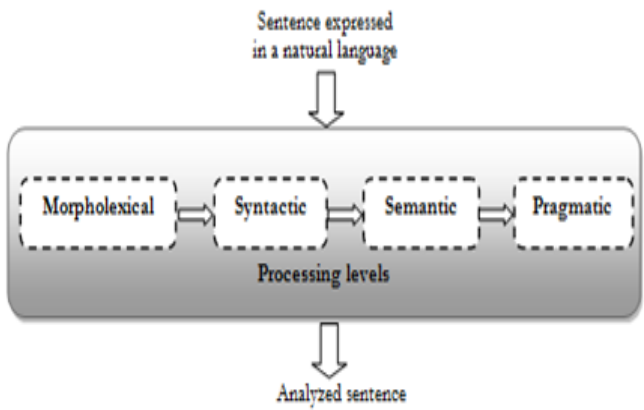

**Figure 2: The processing levels in NLP**

The morpholexical processing level aims to recognize the structure of words.

The syntactic processing level strives for structuring the formal relationships between words of the statement.

The semantic processing level searches for understanding the meaning of individual words of the statement.

The pragmatic procesing looks for contextualizing words by analyzing the meaning in the context.

Different NLP systems implement or all of these tasks and even a combination of some of them [23].

*3.2.2 NLP Tools*

Many NLP tools have been created in research as well as in industry. However, there are already tools that are well recognized for their mastery in NLP namely: GATE (General Architecture for Text Engineering) [24], Open NLP [25], UIMA (Unstructured Information Management Architecture) [26] and IDE (Insight Discoverer Extractor) [27].

Moreover, since the targeted approach is NLP and text mining based, we have argued that we do not need to reinvent the wheel by rebuilding an NLP tool from the scratch. This is why we choose to reuse one of these latters. To do this, we have searched for their own characteristics and assessed them. Table 2 sums up the assessment results and provides comparisons of the four well known tools: GATE, OpenNLP, UIMA and IDE.

**Table 1: Comparison of NLP tools**

|  | **GATE** | **OpenNLP** | **UIMA** | **IDE** |
|---|---|---|---|---|
| **Creation** | 1995 | 1998 | 2001 | 2002 |
| **Licence** | GNU LGPL | LGPL | Apache (since 2005) | Commercial |
| **Input's type** | Text | Text | Text, image audio and video | Text |
| **Supported language(s)** | 8 | Without precision | Many (without precision) | 16 |
| **Programming language(s)** | Java | Java,Python | Java, c++ | Java |
| **Architecture** | Well defined | Absent | Well defined | Absent |
| **Usage** | High | Medium | Medium | Weak |
| **Documentation** | Rich | Medium | Quite rich | Weak |
| **Maturity** | High | Medium | Medium | Weak |
| **Capacity of integration** | Good | Weak (with UIMA) | Good (GATE and OpenNLP) | Not indicated |
| **Performance metrics** | Supported | Not indicated | Not indicated | Not indicated |

The examination of these results reveals that GATE is the most suitable tool since it is open source and very well documented as well as used in research and industry.

Indeed, GATE is an open source and general framework for text engineering which is capable to solve any text processing problem [24]. It also, supports a diversity of formats (doc, pdf, html, xml, rtf, email, etc.) and multilingual data processing using Unicode as its default text encoding.

In order to analyze process patterns, we use the GATE information extraction tool, named: ANNIE [28] acronym for A Nearly-New Information Extraction system.

As it is illustrated in Figure 4, ANNIE corresponds to pipelined components consisting of a Tokeniser, a Gazetteer (system of lexicons), a Sentence Splitter and a Named Entity Transducer.

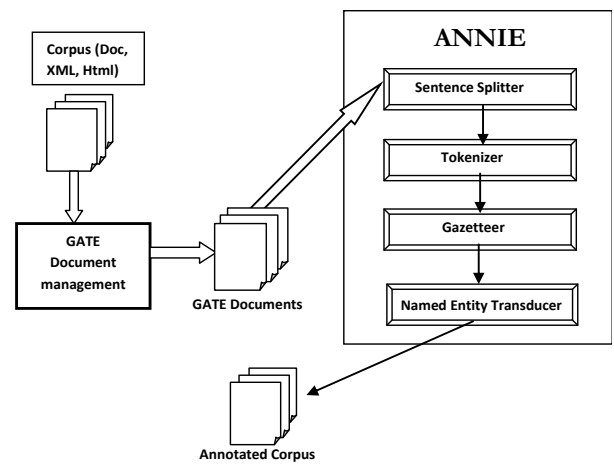

**Figure 3: The ANNIE's Components**

The sentence splitter identifies and annotates the beginning and the end of each sentence. The tokeniser applies basic rules to input text to identify textual objects e.g. punctuation, numbers, symbols and different types. The gazetteer creates annotation to offer information about entities (e.g. persons, organizations, etc.) using lookup lists. The POS tagger produces tags to words or symbols. The Named Entity transducer applies JAPE (Java Annotations Pattern Engine) rules [29] to input text in order to generate new annotations.

## 4. ASAP

Acronym for "Automatic Structuring and Analysis of process Patterns", ASAP aims to improve process patterns reuse by structuring and unifying patterns descriptions.

It is a linguistic approach using performing a NLP technique for the identification of key segments in the descriptions of process patterns, their semantic annotation and then their XML structuring following a unified format.

ASAP comprises two main phases (cf. Figure 4). A first analysis phase consisting in performing lexical, syntactic and semantic analysis of different and unstructered descriptions of process patterns. A second structuring phase converting the analysed patterns to patterns that are semantically annotated following the adopted unification model (cf. Figure 1).

Hence, ASAP consists of an information extraction process from heterogenous and unstructured patterns descriptions and especially a recognition of relevant parts in patterns descriptions and their annotation. So, patterns become described in a unified manner.

In order to reach this goal by implementing the proposed approach, process patterns are analyzed using an extended version of the GATE platform. In fact, to perform the desired annotations, we extended GATE with additional Gazetteer lists as well as additional extraction rules (JAPE rules) to help identify relevant entities in patterns (cf. Figure 5) such as pattern's context, problem, solution, role, etc.

To perform the approch, we considered a corpus of 15 patterns with different formats (Ambler, Gnatz, PROPEL, etc.). Figure 6 illustrates the implementation details of ASAP. The ASAP's implementation is taking place in two main phases: Analysis and Structuring.

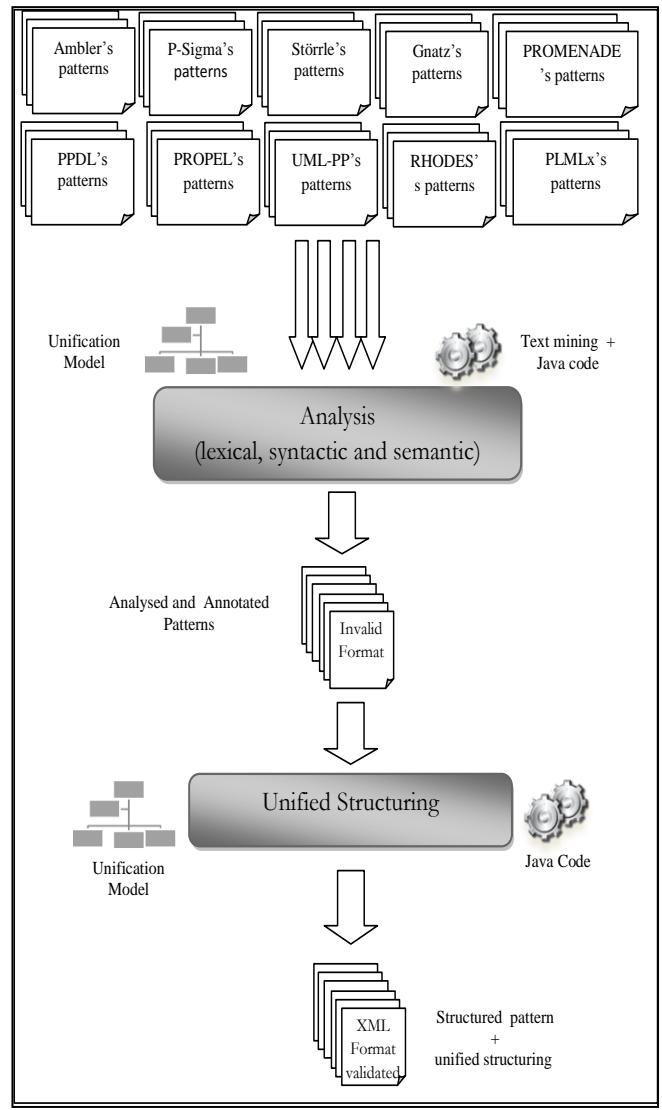

**Figure 4: The proposed ASAP's approach**

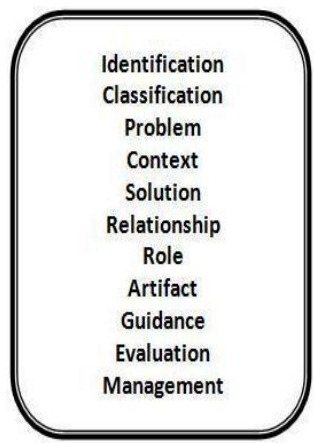

**Figure 5: The added concepts (lists and rules) to ANNIE**

In the first phase of analysis, patterns are analyzed using GATE based on the added concepts and rules. In fact, ANNIE's component begins by recognizing sentences in the processed patterns using the Sentence Splitter component. The sentences are consequently identified using annotations generated by the Sentence Splitter. After that, the Tokeniser splits the text into very simple tokens such as numbers, punctuation and words of different types. Next, Named

Entities are identified in the sentence using annotations such as "Context", "Problem", "Solution", etc. generated from the "Gazetteer Lists" and the "Named Entity Transducer".

The Named Entity Transducer works based on a reference annotation model storing annotations in annotation graphs. A GATE annotation consists of an ID which is unique, a type which denotes the type of the annotation, start and end nodes, and a set of features which provides additional information.

As a result, an XML file (GATE Output) is generated for each pattern provided as input. These XML files involve not only the desired annotations but also other ones that are useless for our purpose, to name just a few, <sentence>, <token> , etc. These annotations should be removed, during the structuring phase.

**Activity**
**Formal solution**
**Intent**
**process**
**Rule**
**Sample execution**
**Semi-formal solution**
**Solution**
**Solution modèle**
**Solution démarche**

**Figure 7: The solution list terminology**

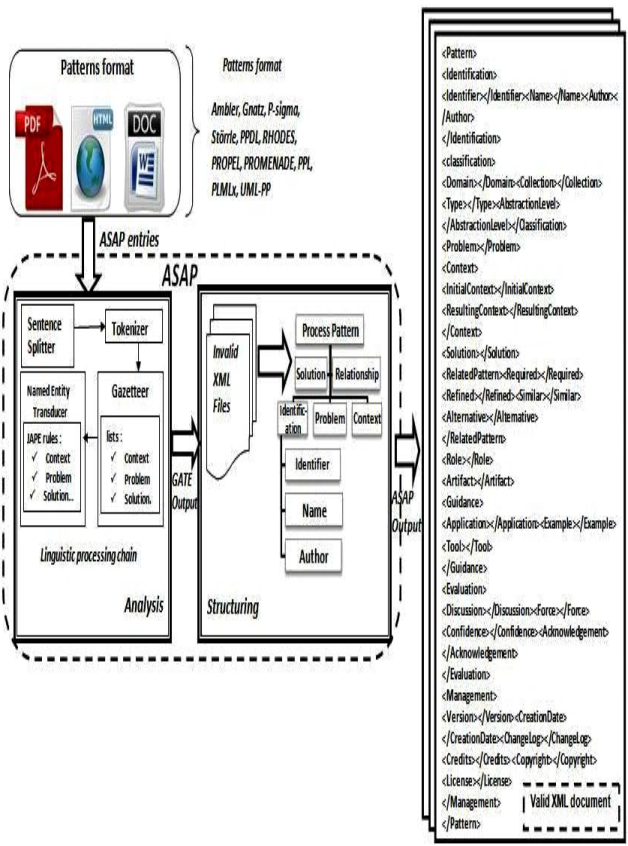

**Figure 6: The implementation details of ASAP**

**Phase: Solution**
Input: Token
Options: control = appelt
**Rule: solution**
(((({Token.string == "Activity"}|{Token.string == "ACTIVITY"}| {Token.string == "activity"}) ({Token.string == " "}| {Token.string == ":"})+)| (({Token.string == "PROCESS"}|{Token.string == "Process"}| {Token.string == "process"})({Token.string == " "}| {Token.string == ":"})+)| (({Token.string == "RULE"}|{Token.string == "Rule"}| {Token.string == ":"})+)| (({Token.string == "SOLUTION"}|{Token.string =="Solution"}| {Token.string == "solution"}|{Token.string == "solution"}) ({Token.string == " "}|{Token.string == ":" })+)| (({Token.string == "SAMPLE"}|{Token.string == "Sample"} ({Token.string == "EXECUTION"}| {Token.string == "execution"}| {Token.string == "execution"})({Token.string == " "}|{Token.string == ":"})+) ((({Token.string == "SEMI"}|{Token.string == "Semi"}| {Token.string == "semi-formal solution"}))) :Solution --> :Solution.Solution = {kind= "Solution", rule= "solution" }

**Figure 8 : The Solution JAPE rule**

JAPE rules: JAPE rules have been added to recognize the terminology used in a given pattern's description and to annotate it in a unified manner . these rules are regrouped into phases such as context phase, solution phase, problem phase, classification phase, relationship phase, identification phase, role phase, artifact phase, guidance phase, evaluation phase and management phase.

Figure 8 shows an excerpt of a JAPE rule identifying a pattern's Solution whose candidate terms are illustrated in Figure 7 i.e. "activity", "intent", "process", "solution", etc.

## 5. EXPERIMENTATIONS

As stated before, we have built our corpus by collecting 15 patterns with different forms and formats. The proposed approach, ASAP, has been implemented with Java programming language and on NetBeans IDE 6.8. The ASAP's system integrates GATE as well as ANNIE as APIs in order to reach the approach goals.

Figure 9 illustrates the experimentation's result of the proposed approach given one process pattern. At the beginning, GATE is being loaded into the ASAP's system and the processing pipeline is performed to generate an XML file

The structuring phase aims to clean and validate XML generated files in order to obtain Valid XML files according to the adopted unified pattern description model. Indeed, during this phase, the ASAP's system should check the integrity of the information obtained from the analysis phase with respect to the grammar used for representing the desired unified format of process patterns.

The GATE extensions that we made concerns :
Gazetteer lists: these lists store the terminologies used to represent pattern's concepts such as: evaluation list, artifact list, classification list, domain list, type list, context list, guidance list, identification list, name list, identifier list, management list, problem list, relationships list, alternative list, similar list, use list, roles list, solution list, author list, abstraction level list, collection list. All these new lists have been successfully integrated and tested on our patterns' corpus. Figure 7 provides an illustration of a gazetteer list representing the pattern's solution.

containing all the targeted annotations. Then, the file is being automatically, cleaned and structured by removing all the unnecessary tags like <sentence>, <token>, via the ASAP's system and according to the unification model.

This speed is justified by the use of the Java programming language and the GATE embedded library as well as ANNIE rather than loading the general GATE platform.

Another kind of evaluation concerns the annotation extraction performance. This latter could be evaluated in terms of three metrics: Precision, Recall and F-measure [30].

The Precision metric measures the number of items correctly identified compared to the number of elements identified. In other words, it measures how many terms were correctly identified by the system. More the precision is close to 1, more identification (annotation) is correct. The Precision is calculated as follows:

$$\text{Precision} = \frac{Correct + \left(\frac{1}{2}\right)*Partial}{correct + false + \left(\frac{1}{2}\right)*partial}$$

The Recall metric measures the number of correctly identified items as a percentage of the total number of correct items. Indeed, it measures how many of the items that should have been identified were identified regardless of how much false identifications were made. Higher the recall, better the system could correctly identify all the elements. Recall is calculated as follows:

$$\text{Recall} = \frac{Correct + \left(\frac{1}{2}\right)*Partial}{Correct + missing + \left(\frac{1}{2}\right)*partial}$$

The F-measure metric combines the precision and recall with weights (β> 0). This measure is calculated as follows:

$$\text{F-mesure} = \frac{(\beta^2+1)*Precision*Recall}{(\beta^2*Recall)+Precision}$$

Each measure is calculated using three different criteria: "strict" "lenient" and "average". The measure "Strict" considers all partially correct answers as incorrect answers. However, the measure "Lenient" considers all partially correct answers as correct answers. The measure "Average" affects half weight to partially correct answers.

In order to measure the annotation extraction performance, GATE provides a tool named AnnotationDiff [31] enabling two sets of annotations in one or two documents to be compared, in order either to compare a system-annotated text with a reference (hand-annotated) text, or to compare the output of two different versions of the system (or two different systems). For each annotation type (e.g. context, problem, solution, relationship, etc.), figures are generated for precision, recall, F-measure. Each of these can be calculated according to 3 different criteria: strict, lenient and average.

To measure the performance of the annotation extraction, we manually identified semantic annotations from a pattern description. Then, using the AnnotationDiff Tool, we compared
The generated set of annotations with the ones extracted through the ASAP's system as depicted in Figure 10.

The key document "patrons.xml" represents the hand annotated document and the response document "patrons.docx" is the ASAP's system one document. So, the AnnotationDiff Tool could compare these two documents annotation by annotation. For example in the Figure 10, the comparison concerns the annotation "Problem".

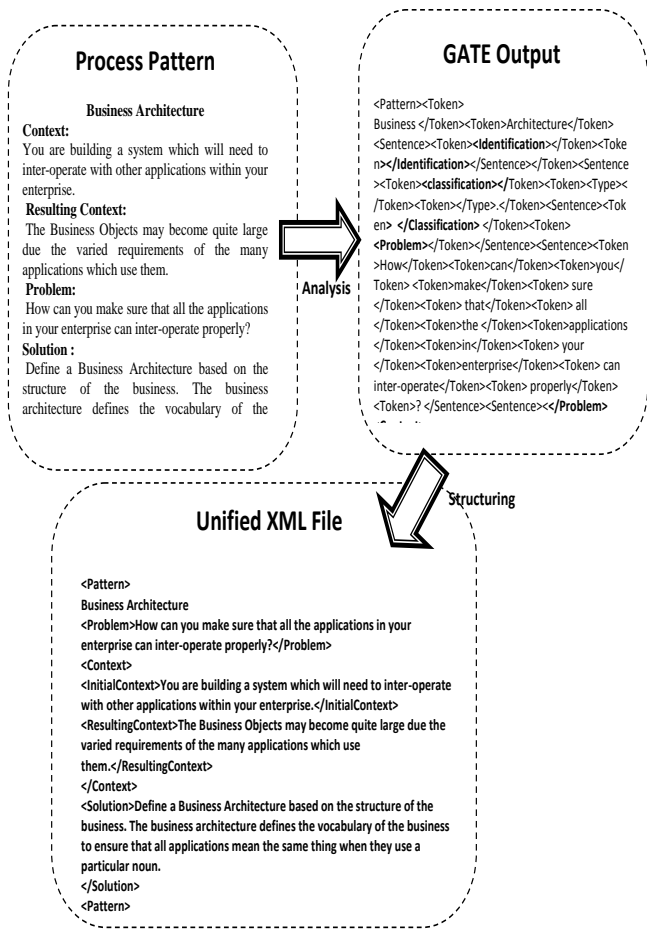

**Figure 9: An illustrative example**

In order to evaluate the ASAP's system performance, we compared it to the GATE framework in term of response time. Table 2 reveals the comparison results for 1, 5, 10 and finally the total number of patterns. In fact, our goal is to evaluate the response time analysis while increasing the size of the corpus.

**Table 2. Response time evaluation**

| Patterns' number | Response time of GATE (in seconds) | Response time of ASAP (in seconds) |
|---|---|---|
| 1 | 6 | 2 |
| 5 | 12 | 4 |
| 10 | 35 | 10 |
| 15 | 40 | 17 |

Given that the response time in ASAP's system presents the response time of analysis and structuring, we notice that the response time is not sensitive to the size of the corpus, for example, a corpus of 10 patterns did not take 10 * 2s (8s is the response time for analyzing a pattern in ASAP) and ASAP is faster than GATE while increasing the size of patterns'corpus.

**Figure 10: Annotation quality evaluation**

The results shown on the left side represent problem annotations extracted by the system and the other ones on the right side concern problem annotations that were manually created. The interpretation of these results, regarding the three introduced metrics, reveals that all annotations are correct (8 correct annotations), Recall, Precision and F-measure measures are always equal to 1, which explains the good performance of our system.

## 6. CONCLUSION

This paper presents a part of our ongoing research work in which we propose a semantic approach for process patterns unification through the automatic analysis and the structuring of their descriptions.

The conducted experimentations show that the approach and its implementation generate high-quality annotations of unstructured and heterogeneous descriptions of process patterns.

The proposed approach ASAP, provides a good starting point as well as a strong foundation for a holistic semantic approach improving process patterns capitalization and reuse [13].

As future work, we aim to extend ASAP by developing a method to automatically convert process patterns provided as XML unified files (ASAP's outputs) to semantic OWL files as ontology's instances.

In addition, we are planning to integrate information extraction possibilities from images (diagrams, figures, tables) that could be achieved using the UIMA java library.

## 7. ACKNOWLEDGMENTS
The authors would like to thank GATE users and developers for their attention and help.